\pdfoutput=1

\documentclass[11pt]{article}

\usepackage[preprint]{acl}

\usepackage{times}
\usepackage{latexsym}

\usepackage[T1]{fontenc}

\usepackage[utf8]{inputenc}

\usepackage{microtype}

\usepackage{inconsolata}

\usepackage{caption}
\usepackage{graphicx}
\usepackage{float}
\usepackage{subfigure}
\usepackage{subcaption}
\usepackage{stfloats}
\usepackage{appendix}

\usepackage{algorithm}
\usepackage{algpseudocode}

\usepackage{amsmath}

\usepackage{booktabs, multirow}
\usepackage{enumitem}
\usepackage{makecell}

\usepackage{color}
\definecolor{green}{RGB}{33, 166, 117}
\definecolor{red}{RGB}{231, 76, 60}
\definecolor{orange}{RGB}{234, 164, 71}
\definecolor{yellow}{RGB}{226, 156, 69}

\usepackage[misc]{ifsym}
\usepackage{footmisc}
\DefineFNsymbols*{newfootnote}{%
    \textasteriskcentered\dag{\textrm{\Letter}}\textdaggerdbl{\ding{73}}\P{**}%
    {\ding{172}}{\ding{173}}{\ding{174}}}
\setfnsymbol{newfootnote}

\newcommand{\framework}{CoEvol}

\title{\textsc{\framework}: Constructing Better Responses for Instruction Finetuning through Multi-Agent Cooperation}

\newcommand*\samethanks[1][\value{footnote}]{\footnotemark[#1]}
\author{
Renhao Li$^{1,2}$\thanks{Equal contribution.}\thanks{Under the Joint Ph.D. Program between UM and SIAT.} \and
Minghuan Tan$^2$\samethanks[1] \and
Derek F. Wong$^1$\thanks{Corresponding author.} \and
Min Yang$^2$\samethanks[3] \\
$^1$ NLP$^2$CT Lab, Department of Computer and Information Science, University of Macau \\
$^2$ Shenzhen Institute of Advanced Technology, Chinese Academy of Sciences \\
\texttt{li.renhao@connect.um.edu.mo,
derekfw@um.edu.mo} \\
\texttt{\{mh.tan,min.yang\}@siat.ac.cn}
}

\begin{document}
\maketitle
\begin{abstract}
In recent years, instruction fine-tuning (IFT) on large language models (LLMs) has garnered considerable attention to enhance model performance on unseen tasks. Attempts have been made on automatic construction and effective selection for IFT data. 
However, we posit that previous methods have not fully harnessed the potential of LLMs for enhancing data quality. The responses within IFT data could be further enhanced by leveraging the capabilities of LLMs themselves.
In this paper, we propose \textsc{{\framework}}, an LLM-based multi-agent cooperation framework for the improvement of responses for instructions. To effectively refine the responses, we develop an iterative framework following a \textit{debate-advise-edit-judge} paradigm. A two-stage multi-agent debate strategy is further devised to ensure the diversity and reliability of editing suggestions within the framework. Empirically, models equipped with \textsc{{\framework}} outperform competitive baselines evaluated by MT-Bench and AlpacaEval, demonstrating its effectiveness in enhancing instruction-following capabilities for LLMs.
\footnote{Code, datasets, and models can be found at \url{https://github.com/lirenhao1997/CoEvol}}

\end{abstract}

\section{Introduction}
Instruction fine-tuning (IFT) is an effective approach for enhancing the performance of language models in zero-shot and few-shot scenarios on previously unseen tasks.
Improving the instruction-following capabilities of large language models (LLMs) has received increasing attentions from the natural language processing (NLP) community~\cite{ouyang2022training, wei2022finetuned, wang-etal-2023-self-instruct, pmlr-v202-longpre23a}. 
Recent research has been focusing on constructing substantial quantities of IFT data with minimal human effort~\cite{wang-etal-2023-self-instruct,honovich-etal-2023-unnatural}, where data construction is highlighted by researchers from multiple perspectives, including diversity, instruction complexity, and the quality of responses to instructions~\cite{liu2024what}. 

To address the issue of easy or moderately difficult human-crafted instructions, several approaches~\cite{wan-etal-2023-explore, zhao-etal-2024-tree, xu2024wizardlm} have been developed to generate instructions of varying complexity levels. By emphasizing both complexity and diversity of the instructions, \citet{lu2024instag} annotate IFT data using advanced LLMs and introduce a complexity-focused diverse sampling method for data selection.
Since LIMA~\cite{zhou2023lima} suggested that the quality of IFT data is more important than its quantity, a series of data selection methods have been proposed, focusing on extracting high-quality samples from existing datasets of uneven distribution of qualities~\cite{li2023one,liu2024what,chen2024alpagasus,xia2024less}. However, we observe that due to the inherent characteristics of causal language modeling, LLMs sometimes fail to deliver the most comprehensive and reasonable answers they can produce. We posit that previous data construction approaches have not fully harnessed the potential of these LLMs. The responses within IFT data could be further refined by leveraging the capabilities of LLMs themselves.

Due to the diversity and complexity of instructions within IFT data, refining the present response is not a trivial task. Consequently, we are attempting to introduce multi-agents to cooperate in this endeavor. Although multi-agent debate (MAD) has been proven effective in answer improvement~\cite{liang2023encouraging,du2023improving} and evaluation~\cite{chan2024chateval} by prompting the diversity of thought, strengthening the divergent thinking of agents cost-effectively remains challenging. We categorize these approaches into two types based on their debate strategies: free debate and predetermined-position debate.
(1) In a free debate, participants freely express their opinions on the topic, and achieving response diversity requires the inclusion of more participants in the debate framework.
(2) In a predetermined-position debate, one debater opposes the other’s views, and a judge decides which side is more persuasive. However, managing the ``tit for tat'' nature of the debate poses a difficulty, while variables like speaking order and statement length can bias the judge’s decision. All these issues make it challenging to refine responses of IFT data through a multi-agent approach.

\begin{figure*}[!htb]
\centering
\includegraphics[width=\linewidth]{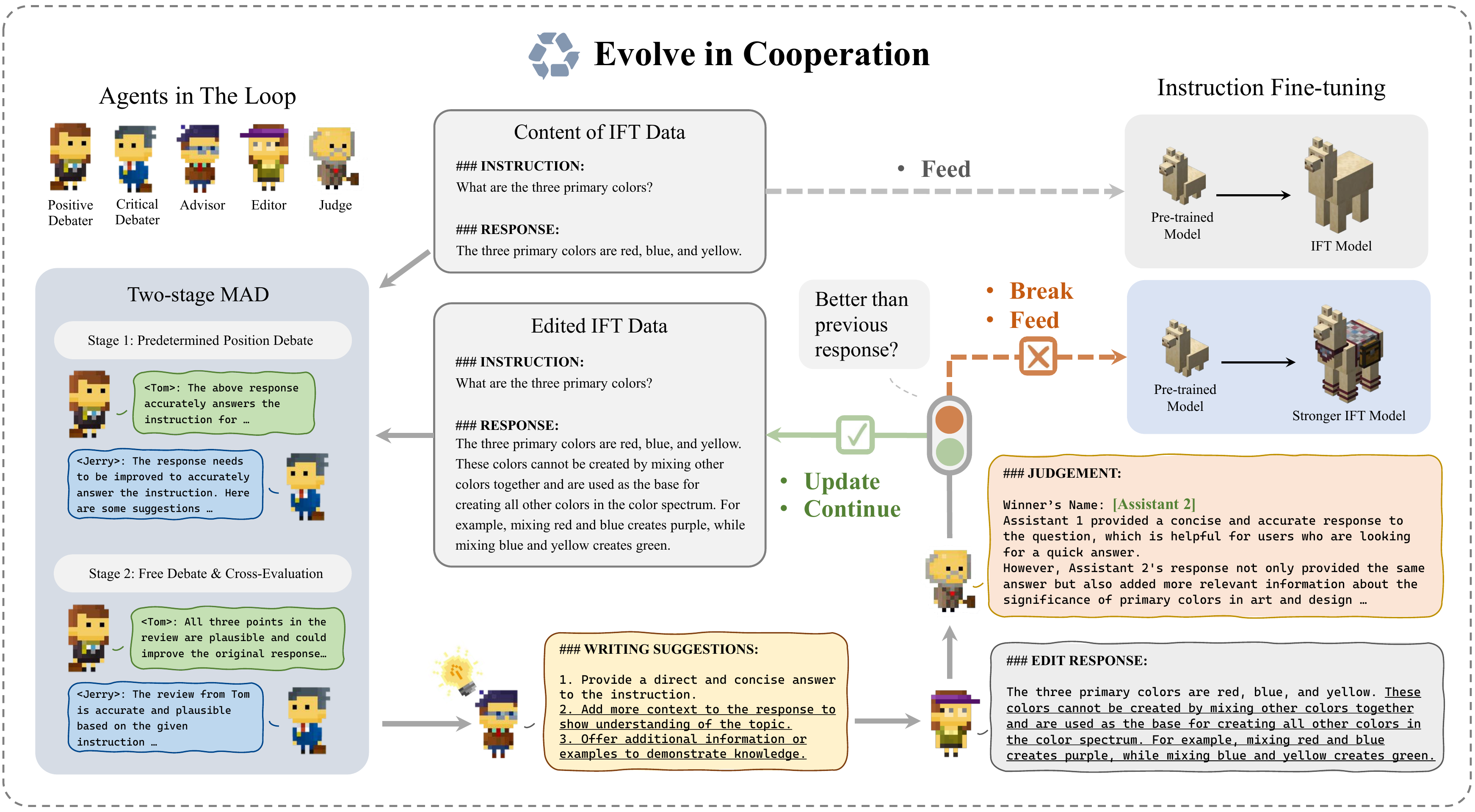}
\caption{Overview of the proposed multi-agent cooperation framework {\framework}.}
\label{fig:framework}
\end{figure*}

In this paper, taking inspiration from human cooperation and competition in real-world society, we endeavor to construct superior responses to instructions with LLM-based multi-agents.
Following a \textit{debate-advise-edit-judge} paradigm, we propose a novel framework named \textsc{{\framework}} to iteratively \underline{evol}ve responses through multi-agents \underline{co}operation. In each iteration, the framework initially enhances the diversity and reliability of subsequent suggestions through a debate between two debaters. The debate history is then presented to an advisor to inform their proposal, followed by an editor who modifies the original response accordingly. 
Finally, a judge evaluates the revised response and determines whether further iteration is necessary. This pipeline enables {\framework} to identify clear and rational evolving directions for the original responses through multi-agent cooperation, resulting in high-quality IFT data.
Furthermore, to address the aforementioned challenges of previous MAD approaches, we design a two-stage strategy within the proposed framework. By combining the advantages of previous methods, our strategy maximizes the diversity of viewpoints while reducing the cost of agents. This novel debate strategy enables the framework to guide responses to evolve in a reliable and trustworthy manner.

We summarize our contributions as follows: 
(1) We introduce a novel {\framework} framework that follows a \textit{debate-advise-edit-judge} paradigm to refine responses within IFT data through multi-agent collaboration. Instead of selecting high-quality data, our approach focuses on editing the low-quality responses of instances to enhance the effectiveness of the IFT data.
(2) We propose a two-stage multi-agent debate strategy designed to maximize the diversity of perspectives within the debate while minimizing the cost of agents.
(3) Empirical results over competitive baselines demonstrate the effectiveness and universality of {\framework}.

\section{Method}
In this section, we introduce our proposed framework for efficiently editing IFT data. Figure~\ref{fig:framework} shows the architecture of {\framework}, where five LLM-based agents: two counter-part debaters, an advisor, an editor, and a judge are assigned in one pipeline to finish the task altogether.

\subsection{Task Assignment}
Before we delve into the detailed design of {\framework}, concepts involved in the IFT task should be clarified first. In this paper, an IFT data sample is denoted by $x$, which comprises the instruction, input, and output components. The instruction typically serves as a description of a task, while the input represents the specific content of this task. When provided with the given instruction and input, the output refers to the response $r$ generated by LLMs.

\subsection{Evolution Pipeline}
We find that while LLMs may not consistently offer comprehensive and exhaustive responses, they excel in recognizing shortcomings within provided responses and offering recommendations for augmentation. Based on this observation, we conceive the idea of integrating LLM-based multi-agents into one pipeline for the iterative refinement of imperfect data samples.

Concretely, we first initialize an experienced advisor $A_{adv}$ through role-play, and ask it to propose writing suggestions for the given data sample $x$:
\begin{equation}
\label{eq:h_adv}
h_{adv} \leftarrow A_{adv}(\hat{x}, t_{adv})
\end{equation}
where $\hat{x}$ denotes the text sequence constructed by filling a given template with the data sample $x$, and $t_{adv}$ signifies the task prompt for the agent advisor.
It merits attention that when the original response is presented to the LLM, it tends to proffer more specific suggestions, whereas when given only the instruction, it provides more general advice which is less helpful for response improvement.
Then we assign an professional editor $A_{edt}$ to modify the original response $r$ referring to the generated suggestions:
\begin{equation}
\label{eq:h_edt}
h_{edt} \leftarrow A_{edt}(\hat{x}, t_{edt}, h_{adv})
\end{equation}
where $t_{edt}$ signifies the task prompt for the agent editor. Denote the edited response as $r'$, a helpful judge $A_{jdg}$ is introduced to compare the helpfulness, relevance, accuracy, and level of details of the original response $r$ and the edited response $r'$:
\begin{equation}
\label{eq:h_jdg}
h_{jdg} \leftarrow A_{jdg}(\hat{x}, t_{jdg}, h_{edt})
\end{equation}
Throughout this process, we discern that the numeric outputs of LLMs occasionally do not correspond with their textual content. For this reason, rather than soliciting the judge to rate responses and compare scores, we instruct it to select the superior response or declare a draw directly from the two presented responses. Moreover, to mitigate the existing position bias inherent within LLM judges~\cite{ko-etal-2020-look,shen-etal-2023-large}, we follow \citet{chen2024alpagasus} to switch the order of original and edited response and make two judgments.

With these judgments, we then respectively calculate scores for $r$ and $r'$ according to the following criteria:
\begin{equation}
\label{eq:s_r}
	s(r)=\left\{
	\begin{aligned}
		1 & , & \text{$r$ is better or tie}\\
		0 & , & \text{otherwise}
	\end{aligned}
	\right.
\end{equation}
Subsequently, these scores will determine whether to continue data evolution or stop the loop:
\begin{itemize}[leftmargin=*]
    \item[-] $s(r') > s(r)$: The edited response $r'$ will be forwarded to the next loop, replacing the original response $r$. In this case, {\framework} will continue to execute the above pipeline.
    \item[-] $s(r') \leq s(r)$: The original response $r$ will be kept as the final response for model fine-tuning. In this case, {\framework} will stop the loop.
\end{itemize}

\subsection{Debate Strategy}
In our framework, the writing suggestions provided by the advisor explicitly determine the direction of response evolution and thus play an important role in the evolution pipeline. To further increase the diversity while ensuring the reliability of these suggestions, we devise a two-stage debate strategy. It combines the advantages of both the predetermined-position debate and the free debate strategy, providing supplemental information from different perspectives to assist the agent advisor in proposing more reliable writing suggestions. More specifically, we employ a predetermined-position debate in the first round, subsequently shifting to a free debate in the second round and conducting a cross-evaluation between the two agents. To mitigate the influence of speaking order on the debate, we allow the debaters to speak concurrently. 

\paragraph{Predetermined-Position Debate.}
In the first round of the debate, we initialize two debate agents with predetermined-positions. To facilitate better engagement from the positive and critical debaters with our prompts, we initially apply role-play prompts to define their respective characters. Subsequently, we provide them with structured sample content, denoted as $\hat{x}$, along with a specific task $t$. This setup helps us acquire their arguments, which are denoted as $g^{pred}_{pos}$ for the positive debater and $g^{pred}_{crt}$ for the critical debater:
\begin{align}
    \label{eq:g_pred_pos}
    g^{pred}_{pos} &\leftarrow A_{pos}(\hat{x}, t^{pred}_{pos}) \\
    \label{eq:g_pred_crt}
    g^{pred}_{crt} &\leftarrow A_{crt}(\hat{x}, t^{pred}_{crt})
\end{align}
\noindent where $A_{pos}$ and $A_{crt}$ respectively refer to the initialized positive debater and negative debater, $t^{pred}_{pos}$ and $t^{pred}_{crt}$ denote the task prompts in predetermined-position debate stage.
To maximize the initial diversity of the debate, we instruct the two debaters with contrary task prompts. Regarding a debate topic ``whether the original response accurately answers the given instruction'', we prompt the positive debater to support the claim and give reasons, while asking the critical debater to argue against it and offer suggestions on how to improve the original response. In this way, we ensure a distinct contrast in viewpoints from the outset of the debate.

\begin{algorithm}[!t]
    \caption{Pseudocode of \framework}
    \label{alg:framework}
    \begin{algorithmic}[1]
    \Require $x$: IFT data sample; $r$: original response; $K$: maximum rounds of evolution;
    \Ensure $x'$: evolved sample
    \State initialize agents $A_{pos}$, $A_{crt}$, $A_{adv}$, $A_{edt}$, and $A_{jdg}$;
    \State $k \leftarrow 1$;
    \While {$k \leq K$}
        \State construct structured $\hat{x}$ from $x$;
        \State generate $g^{pred}_{pos}$ with $A_{pos}$; \Comment{Eq.~\ref{eq:g_pred_pos}}
        \State generate $g^{pred}_{crt}$ with $A_{crt}$; \Comment{Eq.~\ref{eq:g_pred_crt}}
        \State generate $g^{free}_{pos}$ with $A_{pos}$; \Comment{Eq.~\ref{eq:g_free_pos}}
        \State generate $g^{free}_{crt}$ with $A_{crt}$; \Comment{Eq.~\ref{eq:g_free_crt}}
        \State construct debate history $G_{dbt}$;
        
        \State generate $h_{adv}$ with $A_{adv}$;
        \Comment{Eq.~\ref{eq:h_dbt_adv}}
        
        \State generate $h_{edt}$ with $A_{edt}$;
        \State extract edited response $r'$; \Comment{Eq.~\ref{eq:h_edt}}
        
        \State switch order of responses and generate two judgments $h_{jdg}$ with $A_{jdg}$; \Comment{Eq.~\ref{eq:h_jdg}}
        \State calculate scores $s(r)$ and $s(r')$; \Comment{Eq.~\ref{eq:s_r}}
        \If {$s(r') > s(r)$}
            \State update $x$ with $r'$;
            \State refresh the memory of all agents;
        \Else
            \State \textbf{Break}
        \EndIf
        \State $k \leftarrow k+1$;
    \EndWhile
    \State  $x' \leftarrow x$;
    \State \Return $x'$
    \end{algorithmic}
\end{algorithm}
\paragraph{Free Debate and Cross-Evaluation.}
In the second round of the debate, we instruct $A_{pos}$ and $A_{crt}$ to freely express their opinion and do a cross-evaluation regarding to the previous debate topic. Taking debaters' arguments in the first round as reviews towards the given response, we then request both debaters to evaluate the plausibility of the opposing debater's prior review in this stage:
\begin{align}
    \label{eq:g_free_pos}
    g^{free}_{pos} &\leftarrow A_{pos}(\hat{x}, t^{free}_{pos}, g^{pred}_{crt}) \\
    \label{eq:g_free_crt}
    g^{free}_{crt} &\leftarrow A_{crt}(\hat{x}, t^{free}_{crt}, g^{pred}_{pos})
\end{align}
\noindent where $t^{free}_{pos}$ and $t^{free}_{crt}$ respectively denote the task prompts for the positive debater and the critical debater during the free debate stage. With kept memory from the first debate round, we hope both debaters can make objective arguments, resulting in more reliable evaluations.

\subsection{Data Refinement}
Based on the proposed two-stage debate strategy, we obtain viewpoints related to the original response, which are diverse and reliable. Then we sent the generated debate history to the agent advisor $A_{adv}$, asking it to summarize credible ideas from the dialogue and rewrite them into no more than 3 writing suggestions for improving the given response. This process is correspondingly referred to as:
\begin{equation}
\label{eq:h_dbt_adv}
h_{adv} \leftarrow A_{adv}(\hat{x}, t_{adv}, G_{dbt})
\end{equation}
\noindent where $G_{dbt}=\{g^{pred}_{pos}, g^{pred}_{crt}, g^{free}_{pos}, g^{free}_{crt}\}$ denotes the debate history comprising all the arguments presented by both debaters.

To sum up, the proposed framework {\framework} works as follows: For each data sample awaiting improvement, two agent debaters are initially involved in a debate to present their arguments. Then the agent editor raises helpful writing suggestions for response improvement based on the debate history. Referring to these suggestions, an agent editor tries to generate an edited response. Finally, a judge is assigned to compare the original response and the edited response, deciding whether to update the response for further evolution or keep the original one and break the loop. Additionally, a hyperparameter $K$ is set to control the maximum iteration for data evolution. We refresh the memory (session history) of all agents after each iteration.
We present the pseudo-code of {\framework} in Algorithm~\ref{alg:framework}. All prompts, including role-play, task assignment, and template used to regularize context for agents are shown in detail in Appendix~\ref{app:agent_prompts}.

\section{Experimental Results}
\subsection{Preliminary Experiment}
\label{sec:pre_exp}
\begin{table*}[!htp]
\centering
\caption{Results of different instruction-tuned models on MT-Bench and AlpacaEval based on the GPT-4 automatic evaluation. We also show the data source, data construction method, data size, and model alignment method for training. The best result is bolded, while the second-best result is underlined. $\diamondsuit$: results extracted from the official rank list; $\heartsuit$: results reproduced by ourselves.}
\label{tab:pre_exp}
\scalebox{0.81}{
\begin{tabular}{lccccc}\toprule
\textbf{Model} &\textbf{Data Source} &\textbf{Data Construction / Size} &\textbf{Alignment} &\textbf{MT-Bench} &\textbf{AlpacaEval (\%)} \\\midrule
LLaMA2-7B-Chat$^{\diamondsuit}$ &- &- / >100K + 1M &SFT + RLHF &\textbf{6.27} &\textbf{71.4} \\\midrule
Alpaca2-7B$^{\heartsuit}$ &\multirow{4}{*}{$D_{alpaca}$} &Full / 52K &SFT &3.94 &20.15 \\
AlpaGasus2-7B$^{\heartsuit}$ & &Select / 9K &SFT &2.86 &8.38 \\
LLaMA2-7B-SFT$_{random}$ & &Random / 9K &SFT &2.28 &8.31 \\
CoEvol-LLaMA2-7B$_{\textsc{ChatGPT}}$ & &Random + Evol / 9K &SFT &\underline{4.32} &\underline{43.55} \\
\bottomrule
\end{tabular}
}
\end{table*}

\paragraph{Experimental Setup}
In this section, we aim to validate the capability of the proposed framework {\framework} in enhancing the quality of randomly selected IFT data.
We use the 52K Alpaca data~\cite{taori2023stanford} as the base data pool $D_{alpaca}$. Previously, \citet{chen2024alpagasus} select 9K high-quality data from $D_{alpaca}$ according to ratings given by gpt-3.5-turbo (abbreviated as \textit{Select}) and fine-tune a stronger model named \textsc{AlpaGasus}. Deem it as a competitive baseline approach, we first randomly select the same amount of 9K data from $D_{alpaca}$ to form a dataset (abbreviated as \textit{Random}). Then {\framework} is applied to these data for response improvement and obtain a dataset with higher quality (abbreviated as \textit{Random + Evol}). To fully evolve data samples, we set the maximum number of iterations in {\framework} to 3. We fine-tune the pre-trained language model LLaMA2-7B~\cite{touvron2023llama2} on aforementioned datasets respectively and evaluate them on MT-Bench~\cite{zheng2023judging} as well as AlpacaEval~\cite{alpaca_eval} automatically.
We note there exists another relevant IFT data augmentation method~\cite{subramaniam2024debategpt}, where codes and datasets have not been released. Considering the incomplete prompts provided in the paper, we exclude it from the comparative methods in our experiments.
Please note that in this section, we employ ChatGPT (gpt-3.5-turbo-1106)~\cite{openai2022chatgpt} as multi-agents within {\framework}. Detailed setup of data evolution is reported in Appendix~\ref{app:evolve_details}. Settings of fine-tuning are provided in Appendix~\ref{app:train_details}.

\paragraph{Main Results}
Table~\ref{tab:pre_exp} illustrates evaluation results of different LLaMA2-7B-based instruction-tuned models on MT-Bench and AlpacaEval. Since AlpaGasus’s data is not officially disclosed, we reproduce AlpaGasus2-7B on 9K filtered data released by an unofficial implementation.\footnote{\url{https://github.com/gpt4life/alpagasus}}
Among all these models, LLaMA2-7B-Chat performs the best, leveraging SFT and RLHF on massive data. In models fine-tuned with less data, the model trained on data evolved by {\framework} performs better. As shown in the results, {\framework} significantly improves the quality of 9K randomly sampled data, leading to a superior model for instruction following. It also outperforms models fine-tuned on full 52K data and 9K data selected by LLM-based scoring, This serves as compelling evidence of the efficacy of {\framework} in enhancing the quality of IFT data.

Furthermore, it is worth noting that MT-Bench assesses the model's performance using both first-turn and second-turn responses to questions, thereby evaluating the model's multi-turn instruction-following capability. In contrast, the Alpaca dataset follows a single-turn question-answer format, resulting in a disparity between the model's fine-tuning and the evaluation phase. To narrow this gap and explore the potential of {\framework}, we introduce a more diversified experimental setup and conduct further experiments on the framework.

\subsection{Further Experiment}
\label{sec:fur_exp}
\begin{table*}[!htp]\centering
\caption{Results of different instruction-tuned models on MT-Bench and AlpacaEval based on the GPT-4 automatic evaluation. We also show the data source, data construction method, data size, and model alignment method during training. The best result is bolded, while the second-best result is underlined. $\diamondsuit$: results extracted from the official rank list; $\heartsuit$: results reproduced by ourselves.}
\label{tab:fur_exp}
\scalebox{0.8}{
\begin{tabular}{lccccc}\toprule
\textbf{Model} &\textbf{Data Source} &\textbf{Data Construction / Size} &\textbf{Alignment} &\textbf{MT-Bench} &\textbf{AlpacaEval (\%)} \\\midrule
Mistral-7B-Instruct-v0.1$^{\diamondsuit}$ &- &- &- &6.84 &69.65 \\\midrule
\textsc{Deita}-Mistral-7B$^{\heartsuit}$ &\multirow{3}{*}{$D_{single}$} &Select / 6K &SFT &6.26 &65.49 \\
\textsc{\framework}-Mistral-7B$_{\textsc{ChatGPT}}$ & &Select + Evol / 6K &SFT &6.45 &67.04 \\
\textsc{\framework}-Mistral-7B$_{\textsc{Mixtral}}$ & &Select + Evol 
/ 6K &SFT &6.75 &71.43 \\\midrule
\textsc{Deita}-Mistral-7B$^{\heartsuit}$ &\multirow{3}{*}{$D_{multi}$} &Select / 6K &SFT &7.03 &80.08 \\
\textsc{\framework}-Mistral-7B$_{\textsc{ChatGPT}}$ & &Select + Evol / 6K &SFT &\underline{7.16} &\underline{83.54} \\
\textsc{\framework}-Mistral-7B$_{\textsc{Mixtral}}$ & &Select + Evol 
/ 6K &SFT &\textbf{7.22} &\textbf{89.76} \\
\bottomrule
\end{tabular}
}
\end{table*}

\paragraph{Experimental Setup}
In this section, we delve further into the universality and effectiveness of the proposed framework, wherein four questions require answers: (1) Can {\framework} further improve the ``high-quality'' data mined by advanced data selection approaches? (2) Can {\framework} be applied to different forms of data like multi-turn conversations? (3) Can agents in {\framework} work efficiently based on different LLMs? (4) Can evolved data effectively enhance the instruction-following capability of different pre-trained language models?

We conduct a series of experiments to address the precedent issues. For questions (1) and (2), we construct a data pool $D_{single}$ composed of single-turn IFT data samples, and a data pool $D_{multi}$ composed of multi-turn conversations, separately. Concretely, we construct $D_{single}$ with Alpaca, WizardLM (based on Alpaca and ShareGPT)~\cite{xu2024wizardlm}, Dolly~\cite{conover2023free}, and LIMA~\cite{zhou2023lima}; and construct $D_{multi}$ with UltraChat~\cite{ding-etal-2023-enhancing} and ShareGPT~\cite{chiang2023vicuna}. After establishing the two data pools, we employ an efficient IFT data selection method \textsc{Deita}~\cite{liu2024what} to extract 6K high-quality data from each of these pools. Our aspiration is for the selected ``high-quality'' data to serve as the cutting-edge baseline for model fine-tuning. Then we further apply {\framework} on these data to enhance the quality of data through response improvement.
In response to question (3), we independently employ the proprietary model ChatGPT and the open-sourced model Mixtral-8$\times$7B-Instruct-v0.1~\cite{jiang2024mixtral} to power {\framework} for data evolution.
To answer question (4), we adopt the foundation model Mistral-7B-v0.1~\cite{jiang2023mistral} for fine-tuning. Subsequently, we conduct an automatic evaluation using both MT-Bench and AlpacaEval as benchmarks.

\paragraph{Main Results}
Table~\ref{tab:fur_exp} illustrates evaluation results of different Mistral-7B-based instruction-tuned models on MT-Bench and AlpacaEval.
Among all these models, \textsc{\framework}-Mistral-7B$_{\textsc{Mixtral}}$ performs the best, beneficial from the data evolution on multi-turn conversation data. Based on evaluation results, even for selected high-quality data, further improvements in model performance can be achieved using {\framework} for data refinement. Through model comparison, we demonstrate the effectiveness of this framework across different data formats and agents. Compared to preliminary experiments, our framework has also proven effective with different base models. Based on the observations mentioned above, the four questions previously posed have all been resolved, demonstrating the universality and effectiveness of this framework.

\section{Analysis}
\subsection{Ablation Study}
\label{sec:ablation}
To verify the validity of each component within {\framework}, we conduct an ablation study of the framework. Continuing from experiments in Section~\ref{sec:pre_exp}, we evolve the 9K randomly sampled data from $D_{alpaca}$ under the following settings:

\begin{table}[!t]\centering
\caption{Ablation study of the proposed framework {\framework}. Different components of the pipeline are respectively applied to the 9K random sampled Alpaca data and are then utilized for model fine-tuning. We report the scores of MT-Bench and AlpacaEval based on the GPT-4 automatic evaluation. The best result is bolded, while the second-best result is underlined.}
\label{tab:ablation}
\scalebox{0.78}{
\begin{tabular}{lcc}\toprule
\textbf{Model}  &\textbf{MT-Bench} &\textbf{AlpacaEval (\%)} \\\midrule
LLaMA2-7B-SFT$_{random}$ &2.28 &8.31 \\
\quad - edit &4.05 &26.15 \\
\quad - advise (w/o resp) + edit &3.49 &16.38 \\
\quad - advise + edit &4.09 &30.17 \\
\quad - debate + advise + edit &\underline{4.17} &\underline{38.25} \\
\quad - full \textsc{\framework} &\textbf{4.32} &\textbf{43.55} \\
\bottomrule
\end{tabular}
}
\end{table}

\begin{itemize}[leftmargin=*]
    \item[-] \textbf{edit}: We solely utilize an agent editor to respond according to the given instruction and the input context. The model trained on this data serves as a baseline, excluding the influence of other agents in the proposed pipeline.
	\item[-] \textbf{advise (w/o resp) + edit}: We include an agent advisor to propose writing suggestions and an agent editor to generate responses accordingly. Note that in this experiment, we only show the given instructions and inputs to the advisor.
	\item[-] \textbf{advise + edit}: We include an agent advisor to propose suggestions and an agent editor to output the response accordingly. In this experiment, the original response is also shown to the advisor.
    \item[-] \textbf{debate + advise + edit}: Two agent debaters with opposite positions are further included to provide more reliable references for the advisor.
    \item[-] \textbf{full \framework}: The full proposed framework, where agent debaters, advisor, editor, and judge are all involved in the loop. The maximum number of iterations is set to 3.
\end{itemize}

\begin{figure*}[!htp]
\centering
\subfigure[Proportions of data with different rounds of evolution]{
\begin{minipage}[t]{0.48\textwidth}
\includegraphics[width=8cm]{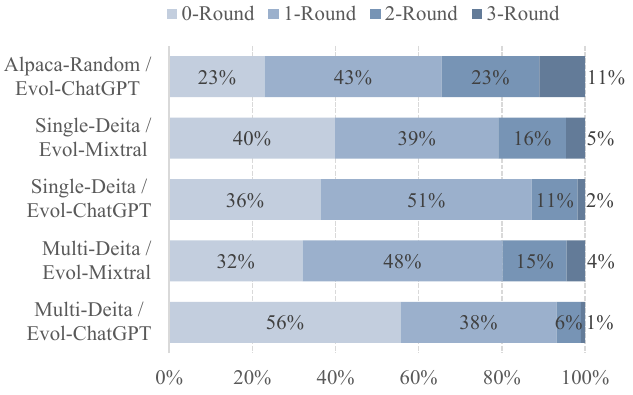}
\label{fig:evol_round}
\end{minipage}
}
\subfigure[Average token lengths of responses]{
\begin{minipage}[t]{0.48\textwidth}
\includegraphics[width=8cm]{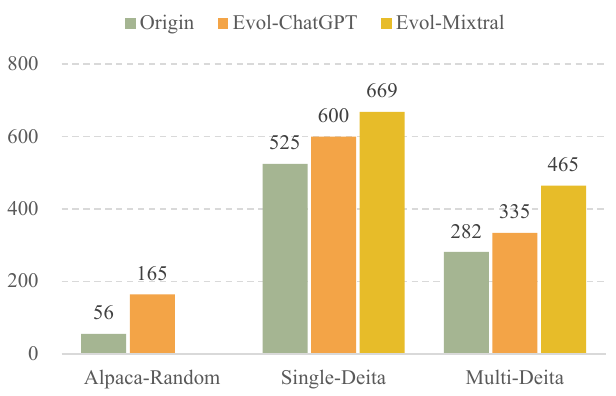}
\label{fig:resp_len}
\end{minipage}
}
\caption{Statistical results of the data evolution process. The proportion of data with different numbers of rounds of evolution driven by {\framework} is shown in figure (a). The average token length of responses in original and evolved data is shown in figure (b). We report statistical results on different datasets and backbone LLMs for agents.}
\end{figure*}

With other settings held constantly, we respectively fine-tune LLaMA2-7B on these evolved data. Evaluation results of the ablated models on MT-Bench and AlpacaEval are shown in Table~\ref{tab:ablation}. According to the results, we observe that the best result is obtained by the complete {\framework}. The participation of each agent enhances the performance of the model, indicating their contribution within the framework. More rounds of data evolution under the guidance of the agent judge also contribute to the data augmentation.
It is worth noting that, prompting agent advisor with only instruction and input compromises the data quality. However, providing the original responses aids the agent advisor in offering more specific suggestions, resulting in an improvement in model performance. We provide more detailed examples of data evolution using ablation models in Appendix~\ref{app:example_of_data_evol}.

\begin{figure}[htb]
\centering
\includegraphics[width=\linewidth]{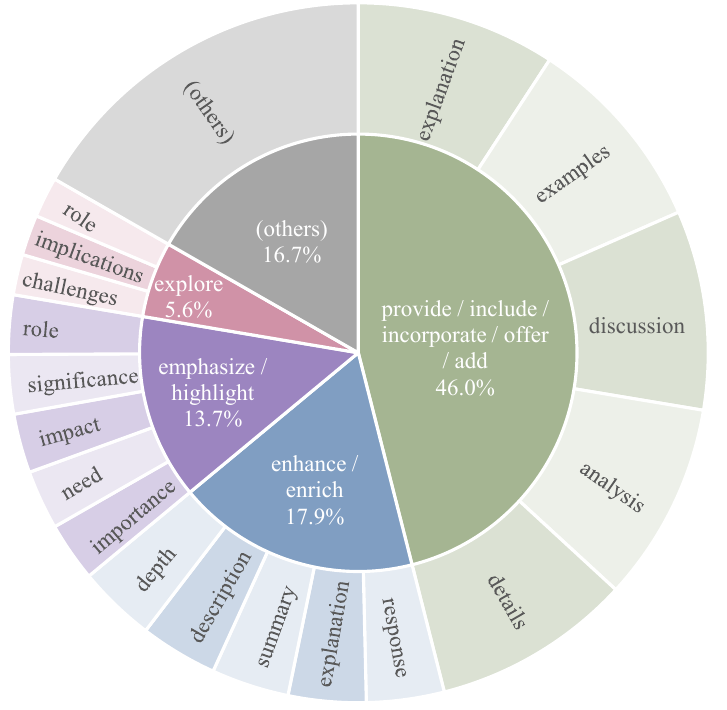}
\caption{Overview of the evolving direction of {\framework}. Due to space constraints, we merge similar root verbs and show parts of verb-object pairs with top counts.}
\label{fig:evol_direct}
\end{figure}

\subsection{Analysis on Data Evolution}
\paragraph{Statistical Results}
To get deep insight into {\framework}, we investigate the data evolution process from multiple perspectives, including statistical results and evolving directions. As illustrated in Figure~\ref{fig:evol_round}, the proportions of data with different rounds of evolution vary among different datasets and backbone LLMs. 
Considering data that has undergone at least one round of evolution, the randomly sampled data from Alpaca dataset has the highest rate of evolution among all these datasets. We attribute this to its poor initial quality of data.
Interestingly, we also observe that {\framework}, when based on Mixtral, has a higher proportion of data that has undergone at least two rounds of evolution compared to that based on ChatGPT.

As longer outputs are preferred by both human and machine evaluators on most datasets~\cite{dubois2023alpacafarm}, the average token length of responses reflects the comprehensiveness of response to some extent. According to Figure~\ref{fig:resp_len}, we observe that the average token length of a response in the evolved dataset has significantly increased compared to the non-evolved dataset.

\begin{table*}[htb]\centering
\footnotesize
\caption{Cases of responses generated by the baseline model \textsc{Deita}-Mistral-7B and proposed model \textsc{\framework}-Mistral-7B$_{\textsc{Mixtral}}$. Instructions are extracted from the AlpacaEval dataset. To highlight the differences between responses, we mark direct responses to instructions in \textcolor{green}{green} and additional relevant information in \textcolor{yellow}{yellow}.}\label{tab:case_study}
\begin{tabular}{p{7.6cm}p{7.6cm}}\toprule
\textbf{Responses from} \textsc{Deita}-Mistral-7B&\textbf{Responses from} \textsc{\framework}-Mistral-7B$_{\textsc{Mixtral}}$\\\midrule
\multicolumn{2}{l}{\textbf{\#1 Instruction: What is the largest star in our galaxy?}}\\
The largest star in our galaxy, the Milky Way, is called VY Canis Majoris ... Its mass is estimated to be around 15-20 times the mass of the Sun ... & The largest star in our galaxy, the Milky Way, is a star known as VY Canis Majoris ... \textcolor{green}{To put that into perspective, if the Sun were the size of a small grape, VY Canis Majoris would be the size of a basketball} ...\\
\midrule
\multicolumn{2}{l}{\textbf{\#2 Instruction: What is the name of the Windows XP background image? Where was it taken and by who?}}\\
The default Windows XP background image is called \"Bliss\" and it was taken by Charles O'Rear. The image was taken in Napa Valley, California, USA. & The Windows XP background image is called \"Bliss.\" It was taken by Charles O'Rear, a professional photographer, in 1996 in Napa Valley, California. \textcolor{yellow}{The image depicts a rolling hillside with a few trees and a blue sky in the background. The image was chosen by Microsoft to represent the serene and optimistic nature of the Windows XP operating system.}\\
\midrule
\multicolumn{2}{l}{\textbf{\#3 Instruction: Write a daily exercise routine for Monday to Sunday with a rest day in between.}}\\
Monday: 30 min cardio, 30 min strength training\newline Tuesday: 45 min yoga, 15 min abs\newline Wednesday: Rest day\newline Thursday: 30 min HIIT, 20 min stretching\newline Friday: 45 min Pilates, 15 min arms\newline Saturday: 60 min outdoor activity (e.g. hiking, biking)\newline Sunday: 30 min cardio, 30 min strength training & Sure, here's a daily exercise routine for Monday to Sunday with a rest day in between:\newline Monday: Start the week with a 30-minute HIIT workout \textcolor{green}{to boost metabolism and burn calories.} \newline Tuesday: Strengthen your core with a 20-minute plank challenge, \textcolor{green}{holding the plank position for as long as possible each day} ... \textcolor{yellow}{Remember to warm up before each workout, stay hydrated, and listen to your body to avoid injury.}\\
\bottomrule
\end{tabular}
\end{table*}

\paragraph{Evolving Directions}
Since the evolving direction of data is determined by suggestions proposed by the agent advisor, we try to investigate which parts of the data are improved by {\framework}. Following \cite{taori2023stanford}, we adopt spaCy\footnote{\url{https://spacy.io/}} to extract the root verb together with its direct object from these suggestions. Figure~\ref{fig:evol_direct} shows an overview of evolving directions of {\framework}. where root verbs and direct objects are extracted from suggestions on 9K randomly sampled Alpaca data. The inner circle of the plot represents the root verb of the suggestions, while the outer circle represents the direct objects. Based on the investigation, a large part of data is evolved by incorporating details and explanations (shown in green), while enriching the existing descriptions (shown in blue) also plays an important role in the process of data augmentation. In addition to these relatively common suggestions, more diverse and specific advice is generated by the framework, yielding a total of 235 root verbs and 4,118 verb-object pairs parsed by the tool.

\subsection{Case Study}
Through specific cases, we compare responses from the baseline model and our proposed model in Table~\ref{tab:case_study} to illustrate the behavior of {\framework}. Enhanced with rich details and examples, the evolved response is more comprehensive and helpful than the original response, without providing content beyond the scope of its instruction.

\section{Related Work}
\label{app:related_work}
\paragraph{Instruction Fine-tuning for LLMs}
In recent years, instruction fine-tuning (IFT) has become a prevalent approach for enhancing the applicability of pre-trained language models (PLMs) and improving their generalization capabilities on unseen tasks~\cite{chung2022scaling, wei2022finetuned}.

For instruction fine-tuning (IFT) on LLMs, attempts~\cite{bach-etal-2022-promptsource, wang-etal-2022-super, zhou2023lima} have been made to construct IFT data with the help of human annotators, yet these methods are time-consuming and labor-intensive. To tackle this issue, pipelines~\cite{wang-etal-2023-self-instruct, honovich-etal-2023-unnatural} are proposed to automatically generate data instances from seed tasks. In addition to single-turn instruction following data~\cite{taori2023stanford, peng2023instruction}, large-scale multi-turn dialogues for IFT are also constructed to further enhance LLMs on chat scenarios~\cite{chiang2023vicuna, xu-etal-2023-baize, ding-etal-2023-enhancing}.
Concurrently, a series of studies show that the complexity, diversity, and quality of IFT data significantly influence model alignment.
Concerning the complexity of instructions, approaches~\cite{wan-etal-2023-explore, zhao-etal-2024-tree, xu2024wizardlm} are designed to create large amounts of instruction data with different levels of complexity.
Aware of both complexity and diversity of instructions, \citet{lu2024instag} tag data with a strong LLM and introduce a complexity-focus diverse sampling method for data selection.
In the pursuit of data with high-quality response, \citet{chen2024alpagasus} propose a data selection strategy that automatically identifies and removes low-quality data. While \citet{li2023quantity} select cherry samples from the original dataset according to their instruction-following difficulty. More recently, \citet{liu2024what} investigate plenty of existing data selection methods and propose approaches for enhanced data measurement and selection.

Diverging from prior work, we focus on how to further improve the quality of responses in IFT data. Through our proposed framework {\framework}, the potential of LLM-based multi-agents is unleashed in collaboration to automatically edit responses, thereby generating high-quality data for fine-tuning superior LLMs.

\paragraph{LLM-based Multi-agent Frameworks}
As the ability of LLMs to reason and follow instructions continues to emerge, wrapping LLMs with additional memories and planning schemes into autonomous agents is catching increasing attention from researchers. These agents are capable of communicating with natural language, memorizing their experiences, and conducting reflections on assigned tasks.

Previously, Generative Agents~\cite{park2023generative} indicate that LLMs are effective in simulating believable human behavior when agents are interactive from perspectives like observation, planning, and reflection.
\citet{li2023camel} focus on how communicative agents may collaborate autonomously to finish tasks. 
\citet{qian2023communicative} construct a virtual chat-powered company for software development based on multi-agents.
\citet{wang2023unleashing} transform a single LLM into a cognitive synergist by engaging in multi-turn self-collaboration with multiple personas.
These practices demonstrate that multi-agent frameworks built over LLMs can effectively solve collaborative tasks.
Considering autonomous agents can behave both cooperatively and competitively, multi-agent debate (MAD) is also attracting researchers to explore.
\citet{du2023improving} find that the quality of responses can be improved through debating over multiple rounds, while \cite{subramaniam2024debategpt} propose an approach to generate IFT data based on this framework. 
\citet{liang2023encouraging} adopt MAD to address the Degeneration-of-Thought issue of reflection-style methods. 
ChatEval~\cite{chan2024chateval} offers an automatic human-mimicking evaluation process on NLG tasks based on multi-agents.

In this paper, we benefit from the diversity of thoughts introduced by MAD and design a two-stage strategy within our proposed multi-agent cooperation framework {\framework}.

\section{Conclusion}
In this paper, we introduce {\framework}, an innovative framework for efficient quality improvement on IFT data through multi-agent cooperation. To fully exploit the potential of LLMs for response editing, we propose a two-stage MAD strategy to maximize the diversity of perspectives within debate while minimizing the cost of agents. Following a debate-advise-edit-judge paradigm, we establish a pipeline to harness the collective power of agents with distinct roles.
Experimental results substantiate the efficacy of our proposed framework, showcasing its superiority in evolving better IFT data through response augmentation. Codes, datasets, and model weights developed in this paper are publicly available.
We hope this work can offer new perspectives and references for the automatic construction of high-quality data.

\section*{Acknowledgements}
This work was supported in part by the National Key Research and Development Program of China (2022YFF0902100),
the Science and Technology Development Fund, Macau SAR (Grant Nos. FDCT/060/2022/AFJ, FDCT/0070/2022/AMJ), the Multi-year Research Grant from the University of Macau (Grant No. MYRG-GRG2023-00006-FST-UMDF), the Research Program of Guangdong Province (Grant No. 2220004002576, EF2023-00090-FST), National Natural Science Foundation of China (62406314, 62376262), China Postdoctoral Science Foundation (2023M733654), and Guangdong Basic and Applied Basic Research Foundation (2023A1515110496).

\section*{Limitations}
In this work, we proposed {\framework}, an LLM-based multi-agent cooperation framework for improving IFT data quality through response enhancement. Although experimental results demonstrate that our framework is viable, there are still some limitations that need to be considered: (1) We build multi-agents on top of the same LLM, where the inherent consistency of the model may lead to the accumulation of bias. Further experiments should be done to investigate the impact of agents based on different LLMs on the proposed framework. (2) Due to time and cost considerations, we conduct our experiments on advanced LLMs like gpt-3.5-turbo and mixtral. We are also wondering how far the most powerful models like GPT-4 and Claude-3 can go when equipped with {\framework}. We plan to explore this in the future.

\bibliography{anthology, custom}

\begin{thebibliography}{43}
\expandafter\ifx\csname natexlab\endcsname\relax\def\natexlab#1{#1}\fi

\bibitem[{Bach et~al.(2022)Bach, Sanh, Yong, Webson, Raffel, Nayak, Sharma, Kim, Bari, Fevry, Alyafeai, Dey, Santilli, Sun, Ben-david, Xu, Chhablani, Wang, Fries, Al-shaibani, Sharma, Thakker, Almubarak, Tang, Radev, Jiang, and Rush}]{bach-etal-2022-promptsource}
Stephen Bach, Victor Sanh, Zheng~Xin Yong, Albert Webson, Colin Raffel, Nihal~V. Nayak, Abheesht Sharma, Taewoon Kim, M~Saiful Bari, Thibault Fevry, Zaid Alyafeai, Manan Dey, Andrea Santilli, Zhiqing Sun, Srulik Ben-david, Canwen Xu, Gunjan Chhablani, Han Wang, Jason Fries, Maged Al-shaibani, Shanya Sharma, Urmish Thakker, Khalid Almubarak, Xiangru Tang, Dragomir Radev, Mike Tian-jian Jiang, and Alexander Rush. 2022.
\newblock \href {https://doi.org/10.18653/v1/2022.acl-demo.9} {{P}rompt{S}ource: An integrated development environment and repository for natural language prompts}.
\newblock In \emph{Proceedings of the 60th Annual Meeting of the Association for Computational Linguistics: System Demonstrations}, pages 93--104, Dublin, Ireland. Association for Computational Linguistics.

\bibitem[{Chan et~al.(2024)Chan, Chen, Su, Yu, Xue, Zhang, Fu, and Liu}]{chan2024chateval}
Chi-Min Chan, Weize Chen, Yusheng Su, Jianxuan Yu, Wei Xue, Shanghang Zhang, Jie Fu, and Zhiyuan Liu. 2024.
\newblock \href {https://openreview.net/forum?id=FQepisCUWu} {Chateval: Towards better {LLM}-based evaluators through multi-agent debate}.
\newblock In \emph{The Twelfth International Conference on Learning Representations}.

\bibitem[{Chen et~al.(2024)Chen, Li, Yan, Wang, Gunaratna, Yadav, Tang, Srinivasan, Zhou, Huang, and Jin}]{chen2024alpagasus}
Lichang Chen, Shiyang Li, Jun Yan, Hai Wang, Kalpa Gunaratna, Vikas Yadav, Zheng Tang, Vijay Srinivasan, Tianyi Zhou, Heng Huang, and Hongxia Jin. 2024.
\newblock \href {https://openreview.net/forum?id=FdVXgSJhvz} {Alpagasus: Training a better alpaca model with fewer data}.
\newblock In \emph{The Twelfth International Conference on Learning Representations}.

\bibitem[{Chiang et~al.(2023)Chiang, Li, Lin, Sheng, Wu, Zhang, Zheng, Zhuang, Zhuang, Gonzalez, Stoica, and Xing}]{chiang2023vicuna}
Wei-Lin Chiang, Zhuohan Li, Zi~Lin, Ying Sheng, Zhanghao Wu, Hao Zhang, Lianmin Zheng, Siyuan Zhuang, Yonghao Zhuang, Joseph~E. Gonzalez, Ion Stoica, and Eric~P. Xing. 2023.
\newblock \href {https://lmsys.org/blog/2023-03-30-vicuna/} {Vicuna: An open-source chatbot impressing gpt-4 with 90\%* chatgpt quality}.

\bibitem[{Chung et~al.(2022)Chung, Hou, Longpre, Zoph, Tay, Fedus, Li, Wang, Dehghani, Brahma et~al.}]{chung2022scaling}
Hyung~Won Chung, Le~Hou, Shayne Longpre, Barret Zoph, Yi~Tay, William Fedus, Yunxuan Li, Xuezhi Wang, Mostafa Dehghani, Siddhartha Brahma, et~al. 2022.
\newblock Scaling instruction-finetuned language models.
\newblock \emph{arXiv preprint arXiv:2210.11416}.

\bibitem[{Conover et~al.(2023)Conover, Hayes, Mathur, Xie, Wan, Shah, Ghodsi, Wendell, Zaharia, and Xin}]{conover2023free}
Mike Conover, Matt Hayes, Ankit Mathur, Jianwei Xie, Jun Wan, Sam Shah, Ali Ghodsi, Patrick Wendell, Matei Zaharia, and Reynold Xin. 2023.
\newblock Free dolly: Introducing the world’s first truly open instruction-tuned llm.
\newblock \emph{Company Blog of Databricks}.

\bibitem[{Ding et~al.(2023)Ding, Chen, Xu, Qin, Hu, Liu, Sun, and Zhou}]{ding-etal-2023-enhancing}
Ning Ding, Yulin Chen, Bokai Xu, Yujia Qin, Shengding Hu, Zhiyuan Liu, Maosong Sun, and Bowen Zhou. 2023.
\newblock \href {https://doi.org/10.18653/v1/2023.emnlp-main.183} {Enhancing chat language models by scaling high-quality instructional conversations}.
\newblock In \emph{Proceedings of the 2023 Conference on Empirical Methods in Natural Language Processing}, pages 3029--3051, Singapore. Association for Computational Linguistics.

\bibitem[{Du et~al.(2023)Du, Li, Torralba, Tenenbaum, and Mordatch}]{du2023improving}
Yilun Du, Shuang Li, Antonio Torralba, Joshua~B Tenenbaum, and Igor Mordatch. 2023.
\newblock Improving factuality and reasoning in language models through multiagent debate.
\newblock \emph{arXiv preprint arXiv:2305.14325}.

\bibitem[{Dubois et~al.(2023)Dubois, Li, Taori, Zhang, Gulrajani, Ba, Guestrin, Liang, and Hashimoto}]{dubois2023alpacafarm}
Yann Dubois, Xuechen Li, Rohan Taori, Tianyi Zhang, Ishaan Gulrajani, Jimmy Ba, Carlos Guestrin, Percy Liang, and Tatsunori Hashimoto. 2023.
\newblock \href {https://openreview.net/forum?id=4hturzLcKX} {Alpacafarm: A simulation framework for methods that learn from human feedback}.
\newblock In \emph{Thirty-seventh Conference on Neural Information Processing Systems}.

\bibitem[{Honovich et~al.(2023)Honovich, Scialom, Levy, and Schick}]{honovich-etal-2023-unnatural}
Or~Honovich, Thomas Scialom, Omer Levy, and Timo Schick. 2023.
\newblock \href {https://doi.org/10.18653/v1/2023.acl-long.806} {Unnatural instructions: Tuning language models with (almost) no human labor}.
\newblock In \emph{Proceedings of the 61st Annual Meeting of the Association for Computational Linguistics (Volume 1: Long Papers)}, pages 14409--14428, Toronto, Canada. Association for Computational Linguistics.

\bibitem[{Jiang et~al.(2023)Jiang, Sablayrolles, Mensch, Bamford, Chaplot, Casas, Bressand, Lengyel, Lample, Saulnier et~al.}]{jiang2023mistral}
Albert~Q Jiang, Alexandre Sablayrolles, Arthur Mensch, Chris Bamford, Devendra~Singh Chaplot, Diego de~las Casas, Florian Bressand, Gianna Lengyel, Guillaume Lample, Lucile Saulnier, et~al. 2023.
\newblock Mistral 7b.
\newblock \emph{arXiv preprint arXiv:2310.06825}.

\bibitem[{Jiang et~al.(2024)Jiang, Sablayrolles, Roux, Mensch, Savary, Bamford, Chaplot, Casas, Hanna, Bressand et~al.}]{jiang2024mixtral}
Albert~Q Jiang, Alexandre Sablayrolles, Antoine Roux, Arthur Mensch, Blanche Savary, Chris Bamford, Devendra~Singh Chaplot, Diego de~las Casas, Emma~Bou Hanna, Florian Bressand, et~al. 2024.
\newblock Mixtral of experts.
\newblock \emph{arXiv preprint arXiv:2401.04088}.

\bibitem[{Ko et~al.(2020)Ko, Lee, Kim, Kim, and Kang}]{ko-etal-2020-look}
Miyoung Ko, Jinhyuk Lee, Hyunjae Kim, Gangwoo Kim, and Jaewoo Kang. 2020.
\newblock \href {https://doi.org/10.18653/v1/2020.emnlp-main.84} {Look at the first sentence: Position bias in question answering}.
\newblock In \emph{Proceedings of the 2020 Conference on Empirical Methods in Natural Language Processing (EMNLP)}, pages 1109--1121, Online. Association for Computational Linguistics.

\bibitem[{Kwon et~al.(2023)Kwon, Li, Zhuang, Sheng, Zheng, Yu, Gonzalez, Zhang, and Stoica}]{kwon2023efficient}
Woosuk Kwon, Zhuohan Li, Siyuan Zhuang, Ying Sheng, Lianmin Zheng, Cody~Hao Yu, Joseph~E. Gonzalez, Hao Zhang, and Ion Stoica. 2023.
\newblock Efficient memory management for large language model serving with pagedattention.
\newblock In \emph{Proceedings of the ACM SIGOPS 29th Symposium on Operating Systems Principles}.

\bibitem[{Li et~al.(2023{\natexlab{a}})Li, Hammoud, Itani, Khizbullin, and Ghanem}]{li2023camel}
Guohao Li, Hasan Abed Al~Kader Hammoud, Hani Itani, Dmitrii Khizbullin, and Bernard Ghanem. 2023{\natexlab{a}}.
\newblock \href {https://openreview.net/forum?id=3IyL2XWDkG} {{CAMEL}: Communicative agents for ''mind'' exploration of large language model society}.
\newblock In \emph{Thirty-seventh Conference on Neural Information Processing Systems}.

\bibitem[{Li et~al.(2023{\natexlab{b}})Li, Zhang, Li, Chen, Chen, Cheng, Wang, Zhou, and Xiao}]{li2023quantity}
Ming Li, Yong Zhang, Zhitao Li, Jiuhai Chen, Lichang Chen, Ning Cheng, Jianzong Wang, Tianyi Zhou, and Jing Xiao. 2023{\natexlab{b}}.
\newblock From quantity to quality: Boosting llm performance with self-guided data selection for instruction tuning.
\newblock \emph{arXiv preprint arXiv:2308.12032}.

\bibitem[{Li et~al.(2023{\natexlab{c}})Li, Zhang, Dubois, Taori, Gulrajani, Guestrin, Liang, and Hashimoto}]{alpaca_eval}
Xuechen Li, Tianyi Zhang, Yann Dubois, Rohan Taori, Ishaan Gulrajani, Carlos Guestrin, Percy Liang, and Tatsunori~B. Hashimoto. 2023{\natexlab{c}}.
\newblock Alpacaeval: An automatic evaluator of instruction-following models.
\newblock \url{https://github.com/tatsu-lab/alpaca_eval}.

\bibitem[{Li et~al.(2023{\natexlab{d}})Li, Hui, Xia, Yang, Yang, Zhang, Si, Liu, Liu, Huang et~al.}]{li2023one}
Yunshui Li, Binyuan Hui, Xiaobo Xia, Jiaxi Yang, Min Yang, Lei Zhang, Shuzheng Si, Junhao Liu, Tongliang Liu, Fei Huang, et~al. 2023{\natexlab{d}}.
\newblock One shot learning as instruction data prospector for large language models.
\newblock \emph{arXiv preprint arXiv:2312.10302}.

\bibitem[{Liang et~al.(2023)Liang, He, Jiao, Wang, Wang, Wang, Yang, Tu, and Shi}]{liang2023encouraging}
Tian Liang, Zhiwei He, Wenxiang Jiao, Xing Wang, Yan Wang, Rui Wang, Yujiu Yang, Zhaopeng Tu, and Shuming Shi. 2023.
\newblock Encouraging divergent thinking in large language models through multi-agent debate.
\newblock \emph{arXiv preprint arXiv:2305.19118}.

\bibitem[{Liu et~al.(2024)Liu, Zeng, He, Jiang, and He}]{liu2024what}
Wei Liu, Weihao Zeng, Keqing He, Yong Jiang, and Junxian He. 2024.
\newblock \href {https://openreview.net/forum?id=BTKAeLqLMw} {What makes good data for alignment? a comprehensive study of automatic data selection in instruction tuning}.
\newblock In \emph{The Twelfth International Conference on Learning Representations}.

\bibitem[{Longpre et~al.(2023)Longpre, Hou, Vu, Webson, Chung, Tay, Zhou, Le, Zoph, Wei, and Roberts}]{pmlr-v202-longpre23a}
Shayne Longpre, Le~Hou, Tu~Vu, Albert Webson, Hyung~Won Chung, Yi~Tay, Denny Zhou, Quoc~V Le, Barret Zoph, Jason Wei, and Adam Roberts. 2023.
\newblock \href {https://proceedings.mlr.press/v202/longpre23a.html} {The flan collection: Designing data and methods for effective instruction tuning}.
\newblock In \emph{Proceedings of the 40th International Conference on Machine Learning}, volume 202 of \emph{Proceedings of Machine Learning Research}, pages 22631--22648. PMLR.

\bibitem[{Lu et~al.(2024)Lu, Yuan, Yuan, Lin, Lin, Tan, Zhou, and Zhou}]{lu2024instag}
Keming Lu, Hongyi Yuan, Zheng Yuan, Runji Lin, Junyang Lin, Chuanqi Tan, Chang Zhou, and Jingren Zhou. 2024.
\newblock \href {https://openreview.net/forum?id=pszewhybU9} {\#instag: Instruction tagging for analyzing supervised fine-tuning of large language models}.
\newblock In \emph{The Twelfth International Conference on Learning Representations}.

\bibitem[{OpenAI(2022)}]{openai2022chatgpt}
OpenAI. 2022.
\newblock \href {https://openai.com/blog/chatgpt/} {Chatgpt: Optimizing language models for dialogue}.
\newblock \emph{OpenAI Blog}.

\bibitem[{Ouyang et~al.(2022)Ouyang, Wu, Jiang, Almeida, Wainwright, Mishkin, Zhang, Agarwal, Slama, Gray, Schulman, Hilton, Kelton, Miller, Simens, Askell, Welinder, Christiano, Leike, and Lowe}]{ouyang2022training}
Long Ouyang, Jeffrey Wu, Xu~Jiang, Diogo Almeida, Carroll Wainwright, Pamela Mishkin, Chong Zhang, Sandhini Agarwal, Katarina Slama, Alex Gray, John Schulman, Jacob Hilton, Fraser Kelton, Luke Miller, Maddie Simens, Amanda Askell, Peter Welinder, Paul Christiano, Jan Leike, and Ryan Lowe. 2022.
\newblock \href {https://openreview.net/forum?id=TG8KACxEON} {Training language models to follow instructions with human feedback}.
\newblock In \emph{Advances in Neural Information Processing Systems}.

\bibitem[{Park et~al.(2023)Park, O'Brien, Cai, Morris, Liang, and Bernstein}]{park2023generative}
Joon~Sung Park, Joseph O'Brien, Carrie~Jun Cai, Meredith~Ringel Morris, Percy Liang, and Michael~S Bernstein. 2023.
\newblock Generative agents: Interactive simulacra of human behavior.
\newblock In \emph{Proceedings of the 36th Annual ACM Symposium on User Interface Software and Technology}, pages 1--22.

\bibitem[{Peng et~al.(2023)Peng, Li, He, Galley, and Gao}]{peng2023instruction}
Baolin Peng, Chunyuan Li, Pengcheng He, Michel Galley, and Jianfeng Gao. 2023.
\newblock Instruction tuning with gpt-4.
\newblock \emph{arXiv preprint arXiv:2304.03277}.

\bibitem[{Qian et~al.(2023)Qian, Cong, Yang, Chen, Su, Xu, Liu, and Sun}]{qian2023communicative}
Chen Qian, Xin Cong, Cheng Yang, Weize Chen, Yusheng Su, Juyuan Xu, Zhiyuan Liu, and Maosong Sun. 2023.
\newblock Communicative agents for software development.
\newblock \emph{arXiv preprint arXiv:2307.07924}.

\bibitem[{Ren et~al.(2021)Ren, Rajbhandari, Aminabadi, Ruwase, Yang, Zhang, Li, and He}]{ren2021zero}
Jie Ren, Samyam Rajbhandari, Reza~Yazdani Aminabadi, Olatunji Ruwase, Shuangyan Yang, Minjia Zhang, Dong Li, and Yuxiong He. 2021.
\newblock $\{$Zero-offload$\}$: Democratizing $\{$billion-scale$\}$ model training.
\newblock In \emph{2021 USENIX Annual Technical Conference (USENIX ATC 21)}, pages 551--564.

\bibitem[{Shen et~al.(2023)Shen, Cheng, Nguyen, You, and Bing}]{shen-etal-2023-large}
Chenhui Shen, Liying Cheng, Xuan-Phi Nguyen, Yang You, and Lidong Bing. 2023.
\newblock \href {https://doi.org/10.18653/v1/2023.findings-emnlp.278} {Large language models are not yet human-level evaluators for abstractive summarization}.
\newblock In \emph{Findings of the Association for Computational Linguistics: EMNLP 2023}, pages 4215--4233, Singapore. Association for Computational Linguistics.

\bibitem[{Subramaniam et~al.(2024)Subramaniam, Torralba, and Li}]{subramaniam2024debategpt}
Vighnesh Subramaniam, Antonio Torralba, and Shuang Li. 2024.
\newblock \href {https://openreview.net/forum?id=ChNy95ovpF} {Debate{GPT}: Fine-tuning large language models with multi-agent debate supervision}.

\bibitem[{Taori et~al.(2023)Taori, Gulrajani, Zhang, Dubois, Li, Guestrin, Liang, and Hashimoto}]{taori2023stanford}
Rohan Taori, Ishaan Gulrajani, Tianyi Zhang, Yann Dubois, Xuechen Li, Carlos Guestrin, Percy Liang, and Tatsunori~B. Hashimoto. 2023.
\newblock Stanford alpaca: An instruction-following llama model.
\newblock \url{https://github.com/tatsu-lab/stanford_alpaca}.

\bibitem[{Touvron et~al.(2023)Touvron, Martin, Stone, Albert, Almahairi, Babaei, Bashlykov, Batra, Bhargava, Bhosale et~al.}]{touvron2023llama2}
Hugo Touvron, Louis Martin, Kevin Stone, Peter Albert, Amjad Almahairi, Yasmine Babaei, Nikolay Bashlykov, Soumya Batra, Prajjwal Bhargava, Shruti Bhosale, et~al. 2023.
\newblock Llama 2: Open foundation and fine-tuned chat models.
\newblock \emph{arXiv preprint arXiv:2307.09288}.

\bibitem[{Wan et~al.(2023)Wan, Huang, Yang, Quan, Bi, and Shi}]{wan-etal-2023-explore}
Fanqi Wan, Xinting Huang, Tao Yang, Xiaojun Quan, Wei Bi, and Shuming Shi. 2023.
\newblock \href {https://doi.org/10.18653/v1/2023.emnlp-main.587} {Explore-instruct: Enhancing domain-specific instruction coverage through active exploration}.
\newblock In \emph{Proceedings of the 2023 Conference on Empirical Methods in Natural Language Processing}, pages 9435--9454, Singapore. Association for Computational Linguistics.

\bibitem[{Wang et~al.(2023{\natexlab{a}})Wang, Kordi, Mishra, Liu, Smith, Khashabi, and Hajishirzi}]{wang-etal-2023-self-instruct}
Yizhong Wang, Yeganeh Kordi, Swaroop Mishra, Alisa Liu, Noah~A. Smith, Daniel Khashabi, and Hannaneh Hajishirzi. 2023{\natexlab{a}}.
\newblock \href {https://doi.org/10.18653/v1/2023.acl-long.754} {Self-instruct: Aligning language models with self-generated instructions}.
\newblock In \emph{Proceedings of the 61st Annual Meeting of the Association for Computational Linguistics (Volume 1: Long Papers)}, pages 13484--13508, Toronto, Canada. Association for Computational Linguistics.

\bibitem[{Wang et~al.(2022)Wang, Mishra, Alipoormolabashi, Kordi, Mirzaei, Naik, Ashok, Dhanasekaran, Arunkumar, Stap, Pathak, Karamanolakis, Lai, Purohit, Mondal, Anderson, Kuznia, Doshi, Pal, Patel, Moradshahi, Parmar, Purohit, Varshney, Kaza, Verma, Puri, Karia, Doshi, Sampat, Mishra, Reddy~A, Patro, Dixit, and Shen}]{wang-etal-2022-super}
Yizhong Wang, Swaroop Mishra, Pegah Alipoormolabashi, Yeganeh Kordi, Amirreza Mirzaei, Atharva Naik, Arjun Ashok, Arut~Selvan Dhanasekaran, Anjana Arunkumar, David Stap, Eshaan Pathak, Giannis Karamanolakis, Haizhi Lai, Ishan Purohit, Ishani Mondal, Jacob Anderson, Kirby Kuznia, Krima Doshi, Kuntal~Kumar Pal, Maitreya Patel, Mehrad Moradshahi, Mihir Parmar, Mirali Purohit, Neeraj Varshney, Phani~Rohitha Kaza, Pulkit Verma, Ravsehaj~Singh Puri, Rushang Karia, Savan Doshi, Shailaja~Keyur Sampat, Siddhartha Mishra, Sujan Reddy~A, Sumanta Patro, Tanay Dixit, and Xudong Shen. 2022.
\newblock \href {https://doi.org/10.18653/v1/2022.emnlp-main.340} {Super-{N}atural{I}nstructions: Generalization via declarative instructions on 1600+ {NLP} tasks}.
\newblock In \emph{Proceedings of the 2022 Conference on Empirical Methods in Natural Language Processing}, pages 5085--5109, Abu Dhabi, United Arab Emirates. Association for Computational Linguistics.

\bibitem[{Wang et~al.(2023{\natexlab{b}})Wang, Mao, Wu, Ge, Wei, and Ji}]{wang2023unleashing}
Zhenhailong Wang, Shaoguang Mao, Wenshan Wu, Tao Ge, Furu Wei, and Heng Ji. 2023{\natexlab{b}}.
\newblock Unleashing the emergent cognitive synergy in large language models: A task-solving agent through multi-persona self-collaboration.
\newblock \emph{arXiv preprint arXiv:2307.05300}.

\bibitem[{Wei et~al.(2022)Wei, Bosma, Zhao, Guu, Yu, Lester, Du, Dai, and Le}]{wei2022finetuned}
Jason Wei, Maarten Bosma, Vincent Zhao, Kelvin Guu, Adams~Wei Yu, Brian Lester, Nan Du, Andrew~M. Dai, and Quoc~V Le. 2022.
\newblock \href {https://openreview.net/forum?id=gEZrGCozdqR} {Finetuned language models are zero-shot learners}.
\newblock In \emph{International Conference on Learning Representations}.

\bibitem[{Xia et~al.(2024)Xia, Malladi, Gururangan, Arora, and Chen}]{xia2024less}
Mengzhou Xia, Sadhika Malladi, Suchin Gururangan, Sanjeev Arora, and Danqi Chen. 2024.
\newblock \href {https://openreview.net/forum?id=Kw3ckB2Kfc} {{LESS}: Selecting influential data for targeted instruction tuning}.
\newblock In \emph{ICLR 2024 Workshop on Navigating and Addressing Data Problems for Foundation Models}.

\bibitem[{Xu et~al.(2024)Xu, Sun, Zheng, Geng, Zhao, Feng, Tao, Lin, and Jiang}]{xu2024wizardlm}
Can Xu, Qingfeng Sun, Kai Zheng, Xiubo Geng, Pu~Zhao, Jiazhan Feng, Chongyang Tao, Qingwei Lin, and Daxin Jiang. 2024.
\newblock \href {https://openreview.net/forum?id=CfXh93NDgH} {Wizard{LM}: Empowering large pre-trained language models to follow complex instructions}.
\newblock In \emph{The Twelfth International Conference on Learning Representations}.

\bibitem[{Xu et~al.(2023)Xu, Guo, Duan, and McAuley}]{xu-etal-2023-baize}
Canwen Xu, Daya Guo, Nan Duan, and Julian McAuley. 2023.
\newblock \href {https://doi.org/10.18653/v1/2023.emnlp-main.385} {Baize: An open-source chat model with parameter-efficient tuning on self-chat data}.
\newblock In \emph{Proceedings of the 2023 Conference on Empirical Methods in Natural Language Processing}, pages 6268--6278, Singapore. Association for Computational Linguistics.

\bibitem[{Zhao et~al.(2024)Zhao, Yu, Hui, Yu, Li, Huang, Zhang, and Li}]{zhao-etal-2024-tree}
Yingxiu Zhao, Bowen Yu, Binyuan Hui, Haiyang Yu, Minghao Li, Fei Huang, Nevin~L. Zhang, and Yongbin Li. 2024.
\newblock \href {https://aclanthology.org/2024.lrec-main.1460} {Tree-instruct: A preliminary study of the intrinsic relationship between complexity and alignment}.
\newblock In \emph{Proceedings of the 2024 Joint International Conference on Computational Linguistics, Language Resources and Evaluation (LREC-COLING 2024)}, pages 16776--16789, Torino, Italia. ELRA and ICCL.

\bibitem[{Zheng et~al.(2023)Zheng, Chiang, Sheng, Zhuang, Wu, Zhuang, Lin, Li, Li, Xing, Zhang, Gonzalez, and Stoica}]{zheng2023judging}
Lianmin Zheng, Wei-Lin Chiang, Ying Sheng, Siyuan Zhuang, Zhanghao Wu, Yonghao Zhuang, Zi~Lin, Zhuohan Li, Dacheng Li, Eric Xing, Hao Zhang, Joseph~E. Gonzalez, and Ion Stoica. 2023.
\newblock \href {https://openreview.net/forum?id=uccHPGDlao} {Judging {LLM}-as-a-judge with {MT}-bench and chatbot arena}.
\newblock In \emph{Thirty-seventh Conference on Neural Information Processing Systems Datasets and Benchmarks Track}.

\bibitem[{Zhou et~al.(2023)Zhou, Liu, Xu, Iyer, Sun, Mao, Ma, Efrat, Yu, YU, Zhang, Ghosh, Lewis, Zettlemoyer, and Levy}]{zhou2023lima}
Chunting Zhou, Pengfei Liu, Puxin Xu, Srini Iyer, Jiao Sun, Yuning Mao, Xuezhe Ma, Avia Efrat, Ping Yu, LILI YU, Susan Zhang, Gargi Ghosh, Mike Lewis, Luke Zettlemoyer, and Omer Levy. 2023.
\newblock \href {https://openreview.net/forum?id=KBMOKmX2he} {{LIMA}: Less is more for alignment}.
\newblock In \emph{Thirty-seventh Conference on Neural Information Processing Systems}.

\end{thebibliography}

\clearpage

\appendix
\section{Prompts of Multi-Agents}
\label{app:agent_prompts}
\subsection{Sample Prompt Templates}
In this paper, we apply a straightforward rule to construct structured samples for agent prompting. Given the varied sources of IFT datasets used in our experiments, the original sample formats were inconsistent. In practice, vacant sample prompts led agent debaters to comment on information shortages, which is not what we expected. To address this mismatch, we implemented two distinct prompt templates for structured sample construction. We present these prompts in Table~\ref{tab:data_template}.
\subsection{Role-play and Task Prompts}
We allocate five agents within the proposed pipeline to enhance the quality of responses. In this process, each agent is equipped with a role-play and task prompt to guide their generation. For models that allow the specification of system prompts, such as ChatGPT, we utilize the role-play prompts as their system prompts. Conversely, for models that do not support system prompts, such as Mixtral, we incorporate the role-play prompt before the task prompt and provide instructions as a unified input. We present these prompts in Table~\ref{tab:agent_prompt1}, Table~\ref{tab:agent_prompt2}, and Table~\ref{tab:agent_prompt3} for reference.

\section{Details of Data Evolution}
\label{app:evolve_details}
To guarantee the stability of the proposed pipeline, we set several hyperparameters to control the framework. In our experiments, we set the maximum number of iterations for data evolution to 3. For LLM-based agents, the maximum generated tokens are restricted to 1000, the temperature is maintained at 0 for reproducibility, and the top\_p value is set to 1.0. Regarding the data evolution on multi-turn conversations, the expansion of conversation rounds leads to a cumulative increase in historical information within the instructions targeted for optimization. To mitigate resource consumption, we only retain the most recent 3 rounds of conversation for data refinement in the present turn.

\section{Details of Model Training}
\label{app:train_details}
In our experiments, we fine-tune two pre-trained language models on different IFT data. To facilitate model training, we employ DeepSpeed Zero-Stage 2~\cite{ren2021zero} in model fine-tuning.
Specifically, we follow~\cite{chen2024alpagasus} to train LLaMA2-7B with a batch size of 512 over 3 epochs. The learning rate is set to 2e-5, and a cosine warmup scheduler with a warm ratio of 0.1 is employed. During fine-tuning, we utilize the LLaMA2-style template to concatenate queries and responses within multi-turn conversations. The maximum input length for the model is set to 4096.
Regarding Mistral-7B, we follow~\cite{liu2024what} to train the model with a batch size of 512 over 6 epochs. The learning rate is set to 2e-5, and a cosine warmup scheduler with a warm ratio of 0.1 is employed. During fine-tuning, we utilize the Mistral-style template to concatenate queries and responses within multi-turn conversations. The maximum input length for the model is set to 8192.

\section{Task-specific Evaluation}
In Table~\ref{tab:task_eval}, we provide evaluation results of fine-tuned models on different tasks in MT-Bench. According to the results, the improvement of single-turn responses is more significant than the multi-turn responses. Regarding specific tasks, the model trained with evolved data is more reliable on literal tasks such as \textit{humanities} and \textit{roleplay}.

\section{Examples of Data Evolution}
\label{app:example_of_data_evol}
In Table~\ref{tab:example_of_data_evol1} and \ref{tab:example_of_data_evol2}, we show detailed examples of data evolution using {\framework} under different framework settings according to Section~\ref{sec:ablation}, including the original sample, evolving directions in the form of suggestions, and the evolved responses.

\section{Human Evaluation}
To further conduct a human evaluation of this work, we randomly sample 50 cases from the AlpacaEval dataset and generate responses for each instruction using both the baseline model (\textsc{Deita}-Mistral-7B) and the enhanced model (\textsc{\framework}-Mistral-7B$_{\textsc{Mixtral}}$). Following the human evaluation process of~\citet{zhou2023lima}, we engage human annotators to manually score these responses. Together with the evaluation from a GPT-4 LLM judge, we present the comparison results in the form of wins/ties/losses in Table~\ref{tab:huamn_eval}.

\begin{table}[!t]\centering
\caption{Evaluation results from both human and LLM judges comparing the enhanced model (\textsc{\framework}-Mistral-7B$_{\textsc{Mixtral}}$) with the baseline model (\textsc{Deita}-Mistral-7B). ``Win'' means the response generated by the enhanced model is better than the baseline model.}
\label{tab:huamn_eval}
\scalebox{0.95}{
\begin{tabular}{lcccc}\toprule
\textbf{Evaluator}  &\textbf{Win} &\textbf{Tie} &\textbf{Loss} &\textbf{Total} \\\midrule
Human &14 &28 &8 &50 \\
LLM &12 &33 &5 &50 \\
\bottomrule
\end{tabular}
}
\end{table}

\section{Computational Cost}
Although the two-phase MAD strategy is designed to minimize the cost of agent calls, the iterative nature of the proposed framework still makes it costly. In practice, we deploy Mixtral with vLLM~\cite{kwon2023efficient} on 4 * 80GB A100 GPUs as the backbone LLM for {\framework}. It takes approximately 30 hours to refine 6K single-turn IFT data. Concerning multi-turn data, the rounds to be optimized (or the maximum optimized length) correspondingly increase the cost of agent calls. We believe that future development in open-source communities will reduce the cost of obtaining viable agent calls, thereby enhancing {\framework}'s feasibility in practice.

\begin{table*}[!p]\centering
\caption{Prompt template for structured data construction.}\label{tab:data_template}
\begin{tabular}{p{7.7cm}p{7.7cm}}\toprule
\textit{Prompt Template with Given Input} &\textit{Prompt Template without Given Input} \\\midrule
\#\#\# Instruction:\newline
\textcolor{blue}{\{instruction\}}\newline\newline
\#\#\# Input:\newline
\textcolor{blue}{\{input\}}\newline\newline
\#\#\# Response:\newline
\textcolor{blue}{\{output\}}
&\#\#\# Instruction:\newline
\textcolor{blue}{\{instruction\}}\newline\newline
\#\#\# Response:\newline
\textcolor{blue}{\{output\}}
\\
\bottomrule
\end{tabular}
\end{table*}

\begin{table*}[!p]\centering
\caption{Prompts used for two agent debaters.}\label{tab:agent_prompt1}
\begin{tabular}{p{7.7cm}p{7.7cm}}\toprule
\textit{Prompt for Agent Positive Debater} &\textit{Prompt for Agent Critical Debater} \\\midrule
\textbf{Role-Play Prompt} &\textbf{Role-Play Prompt} \\
You are an optimistic person who embodies a mindset that looks for the best in every situation, maintains a positive attitude, and embraces challenges as opportunities for growth and success.
&You are a critical person who tends to view things through critical thinking and provide feedback for improvement or identify areas of concern. \\
\textbf{Task Assignment Prompt (First Round)} &\textbf{Task Assignment Prompt (First Round)} \\
\textcolor{blue}{\{sample\}}\newline
In your opinion, the above response accurately answers the instruction and the input. Please state reasons why the response is accurate if it is used for supervised fine-tuning. &\textcolor{blue}{\{sample\}}\newline
In your opinion, the above response does not accurately answer the instruction and the input. Please offer suggestions on how to improve the response if it is used for supervised fine-tuning. \\
\textbf{Task Assignment Prompt (Second Round)} &\textbf{Task Assignment Prompt (Second Round)} \\
\#\#\# Review from others:\newline
\textcolor{blue}{\{crt\_pred\}}\newline\newline
Above is another review from others, please evaluate the plausibility of each point according to the given instruction and input.
&\#\#\# Review from others:\newline
\textcolor{blue}{\{pos\_pred\}}\newline\newline
Above is another review from others, please evaluate the plausibility of each point according to the given instruction and input. \\
\bottomrule
\end{tabular}
\end{table*}

\begin{table*}[!p]\centering
\caption{Prompts used for agent advisor and agent editor.}\label{tab:agent_prompt2}
\begin{tabular}{p{7.7cm}p{7.7cm}}\toprule
\textit{Prompt for Agent Advisor} &\textit{Prompt for Agent Editor } \\\midrule
\textbf{Role-Play Prompt} &\textbf{Role-Play Prompt} \\
You are an experienced advisor who possesses a high level of expertise in summarizing and giving advice.
&You are a professional editor who possesses a high level of expertise in refining and improving writing content. \\
\textbf{Task Assignment Prompt} &\textbf{Task Assignment Prompt} \\
Below is an instruction that describes a task, paired with an input that provides further context.\newline
\textcolor{blue}{\{sample\}}\newline\newline
The following is a discussion about the given request and response by two reviewers.\newline\newline
\#\#\# Reviewer 1:\newline
\textcolor{blue}{\{pos\_pred\}}\newline\newline
\#\#\# Reviewer 2:\newline
\textcolor{blue}{\{crt\_pred\}}\newline\newline
\#\#\# Reviewer 1:\newline
\textcolor{blue}{\{pos\_free\}}\newline\newline
\#\#\# Reviewer 2:\newline
\textcolor{blue}{\{crt\_free\}}\newline\newline
Extract and summarize credible ideas from the above dialogue and rewrite them into no more than 3 writing suggestions for improving the given response. Directly output these suggestions in separate lines without any foreword or explanation.
&\#\#\# Writing Suggestions:\newline
\textcolor{blue}{\{adv\_sugg\}}\newline\newline
\#\#\# Previous Response:\newline
\textcolor{blue}{\{pre\_resp\}}\newline\newline
Below is an instruction that describes a task, paired with an input that provides further context.\newline
\textcolor{blue}{\{sample\}}\newline\newline
Referring to the above writing suggestions (MUST ignore suggestions beyond your capabilities), modify the previous response and make sure that it appropriately completes the request.\newline
\textcolor{blue}{\{sample\_request\}}\newline\newline
\#\#\# Response:\\
\bottomrule
\end{tabular}
\end{table*}

\begin{table*}[!p]\centering
\caption{Prompts used for the agent judge.}\label{tab:agent_prompt3}
\begin{tabular}{p{7.7cm}p{7.7cm}}\toprule
\textit{Prompt for Agent Judge} &\textit{Prompt for Agent Judge (In Reverse Order)} \\\midrule
\textbf{Role-Play Prompt} &\textbf{Role-Play Prompt} \\
You are a helpful and precise assistant for checking the quality of the response.
&You are a helpful and precise assistant for checking the quality of the response. \\
\textbf{Task Assignment Prompt} &\textbf{Task Assignment Prompt} \\
Below is an instruction that describes a task, paired with an input that provides further context.\newline
\textcolor{blue}{\{sample\_request\}}\newline\newline
[The Start of Assistant 1's Response]\newline
\textcolor{blue}{\{pre\_resp\}}\newline
[The End of Assistant 1's Response]\newline\newline
[The Start of Assistant 2's Response]\newline
\textcolor{blue}{\{new\_resp\}}\newline
[The End of Assistant 2's Response]\newline\newline
[System]\newline
We would like to request your comparison of the performance of two AI assistants in response to the user request displayed above.
Please compare the helpfulness, relevance, accuracy, and level of detail of their responses.
Please first output a single line containing a name indicating whose response is better, <assistant 1> or <assistant 2> or <equal>. In the subsequent line, please provide a comprehensive explanation of your comparison, avoiding any potential bias and ensuring that the order in which the responses were presented does not affect your judgment.
&Below is an instruction that describes a task, paired with an input that provides further context.\newline
\textcolor{blue}{\{sample\_request\}}\newline\newline
[The Start of Assistant 1's Response]\newline
\textcolor{blue}{\{new\_resp\}}\newline
[The End of Assistant 1's Response]\newline\newline
[The Start of Assistant 2's Response]\newline
\textcolor{blue}{\{pre\_resp\}}\newline
[The End of Assistant 2's Response]\newline\newline
[System]\newline
We would like to request your comparison of the performance of two AI assistants in response to the user request displayed above.
Please compare the helpfulness, relevance, accuracy, and level of detail of their responses.
Please first output a single line containing a name indicating whose response is better, <assistant 1> or <assistant 2> or <equal>. In the subsequent line, please provide a comprehensive explanation of your comparison, avoiding any potential bias and ensuring that the order in which the responses were presented does not affect your judgment.\\
\bottomrule
\end{tabular}
\end{table*}

\begin{table*}[!htp]\centering
\caption{Evaluation results of fine-tuned models on different tasks in MT-Bench.}
\label{tab:task_eval}
\scalebox{0.84}{
\begin{tabular}{lccccccccc}\toprule
\textbf{Model} &\textbf{STEM} &\textbf{Writing} &\textbf{Humanities} &\textbf{Extraction} &\textbf{Roleplay} &\textbf{Math} &\textbf{Reasoning} &\textbf{Code} &\textbf{Avg.} \\\midrule
\textsc{Deita}-Mistral-7B &&&&&&&&&\\
\quad - single-turn &8.70 &9.30 &9.50 &7.60 &7.80 &4.40 &5.70 &5.80 &7.35\\
\quad - multi-turn &8.60 &7.40 &9.50 &7.40 &8.50 &2.50 &5.20 &4.60 &6.71\\
\midrule
\textsc{{\framework}}-Mistral-7B &&&&&&&&&\\
\quad - single-turn &9.00 &9.40 &9.70 &7.70 &9.50 &4.60 &5.00 &6.90 &7.73\\
\quad - multi-turn &8.25 &7.70 &9.80 &6.90 &8.40 &2.50 &5.11 &5.00 &6.73\\
\bottomrule
\end{tabular}
}
\end{table*}

\begin{table*}[!p]\centering
\caption{Examples of data evolution using {\framework} under different framework settings (Section~\ref{sec:ablation}). For an original sample from Alpaca dataset, we show suggestions and refined responses given by ablated frameworks.}
\label{tab:example_of_data_evol1}
\begin{tabular}{p{6.0cm}p{9.4cm}}\toprule
\multicolumn{2}{c}{\textbf{Original Sample}} \\\midrule
\textit{\#\# Instruction}\newline How might someone use machine learning to identify fake news? &\textit{\#\# Response}\newline Machine learning could be used to identify fake news by analyzing the content of articles and detecting patterns of deception. The model could take into account factors such as sentence structure, language, and source to identify potentially false information. Additionally, it could also use natural language processing to automatically detect signs of deception in the text. \\\midrule
\multicolumn{2}{c}{\textbf{Edit}} \\\midrule
\textit{\#\# Suggestions}\newline (None) &\textit{\#\# Response}\newline One way to use machine learning to identify fake news is by training a model on a large dataset of both real and fake news articles. The model can then analyze various features of the articles, such as the language used, the sources cited, and the tone of the writing, to identify patterns that are indicative of fake news. By continuously updating and refining the model based on new data, it can become increasingly accurate in distinguishing between real and fake news articles. Additionally, natural language processing techniques can be employed to analyze the content of the articles and detect any misleading or false information. This approach can help in automating the process of identifying fake news and flagging potentially deceptive content for further review. \\\midrule
\multicolumn{2}{c}{\textbf{Advise (w/o resp) + Edit}} \\\midrule
\textit{\#\# Suggestions}\newline 1. Start by researching and understanding the common characteristics and patterns of fake news, such as sensationalism, misleading headlines, and biased sources.\newline
2. Consider using natural language processing techniques to analyze the language and sentiment of news articles, as well as to detect any inconsistencies or contradictions within the content.\newline
3. Utilize supervised learning algorithms to train a model on a labeled dataset of both real and fake news articles, and then use this model to classify new articles as either real or fake based on their features.
&\textit{\#\# Response}\newline To identify fake news using machine learning, one approach is to start by researching and understanding the common characteristics and patterns of fake news, such as sensationalism, misleading headlines, and biased sources. Then, consider using natural language processing techniques to analyze the language and sentiment of news articles, as well as to detect any inconsistencies or contradictions within the content. Additionally, utilizing supervised learning algorithms to train a model on a labeled dataset of both real and fake news articles can be effective. This model can then be used to classify new articles as either real or fake based on their features. \\
\bottomrule
\end{tabular}
\end{table*}

\begin{table*}[!p]\centering
\caption{Examples of data evolution using {\framework} under different framework settings (\textit{continued}).}\label{tab:example_of_data_evol2}
\begin{tabular}{p{6.0cm}p{9.4cm}}\toprule
\multicolumn{2}{c}{\textbf{Advise + Edit}} \\\midrule
\textit{\#\# Suggestions}\newline 1. Consider providing specific examples of machine learning algorithms or techniques that could be used for identifying fake news, such as sentiment analysis or topic modeling.\newline2. Expand on the potential challenges or limitations of using machine learning to identify fake news, such as the need for large, diverse datasets and the potential for bias in the training data.\newline3. Include information on the importance of fact-checking and human oversight in conjunction with machine learning for more accurate identification of fake news. &\textit{\#\# Response}\newline Machine learning can be utilized to identify fake news through various algorithms and techniques. For instance, sentiment analysis can be employed to gauge the emotional tone of an article, helping to detect potential bias or manipulation. Additionally, topic modeling can be used to categorize and analyze the content of news articles, enabling the identification of patterns or inconsistencies that may indicate false information.\newline However, it is important to acknowledge the potential challenges and limitations of using machine learning for this purpose. One such challenge is the need for large and diverse datasets to train the model effectively, as fake news can take many different forms and may be context-dependent. Moreover, there is a risk of bias in the training data, which could lead to inaccurate or skewed results.\newline It is crucial to emphasize that while machine learning can be a valuable tool in identifying fake news, it should be complemented by human oversight and fact-checking. Human judgment and critical thinking are essential for verifying the accuracy of information, and they can help mitigate the limitations and potential biases of machine learning algorithms. \\\midrule
\multicolumn{2}{c}{\textbf{Debate + Advise + Edit}} \\\midrule
\textit{\#\# Suggestions}\newline 1. Include specific details and examples of machine learning algorithms that could be used for identifying fake news, such as supervised learning, unsupervised learning, or deep learning.\newline 2. Discuss specific features or indicators that could be used to train the model, such as linguistic patterns, sentiment analysis, or credibility of sources, to provide a more comprehensive understanding of the factors contributing to the effectiveness of machine learning in identifying fake news.\newline3. Incorporate examples of successful applications of machine learning in identifying fake news to illustrate the practical implementation and impact of machine learning in this domain. Additionally, address potential challenges or limitations in using machine learning for this purpose, such as the need for large and diverse training datasets and the potential for bias in the algorithms.
&\textit{\#\# Response}\newline Machine learning can be utilized to identify fake news by employing various algorithms such as supervised learning, unsupervised learning, or deep learning. For instance, supervised learning algorithms can be trained on labeled datasets of fake and real news articles to classify new articles. Unsupervised learning algorithms can identify patterns and anomalies in the data, while deep learning algorithms can process large amounts of unstructured data to detect fake news.\newline To train the model, specific features and indicators can be used, such as linguistic patterns, sentiment analysis, and credibility of sources. Linguistic patterns can include the use of sensational language or inconsistent information within the article. Sentiment analysis can assess the emotional tone of the content, while evaluating the credibility of sources can involve analyzing the reputation and history of the publishing platform.\newline Successful applications of machine learning in identifying fake news include platforms that use natural language processing to analyze news articles and social media posts to detect misinformation. However, challenges in using machine learning for this purpose include the need for large and diverse training datasets to ensure the model's accuracy and the potential for bias in the algorithms, which may inadvertently label legitimate news as fake based on certain patterns or sources. \\
\bottomrule
\end{tabular}
\end{table*}


\end{document}